\documentclass[authoryear,3p,preprint,review,11pt]{elsarticle}

\usepackage{amssymb}
\usepackage{amsmath}
\usepackage{booktabs}
\usepackage{threeparttable}

\usepackage{xcolor}

\usepackage{ulem}

\makeatletter
\def\ps@pprintTitle{%
 \let\@oddhead\@empty
 \let\@evenhead\@empty
 \def\@oddfoot{\textit{\small{Computers \& Operations Research}}\hfill}%
 \let\@evenfoot\@oddfoot
}
\makeatother

\begin{document}

\begin{frontmatter}



\title{Graph Neural Networks for Job Shop Scheduling Problems: A Survey}

\author[1,2]{Igor G. Smit*} 
\ead{i.g.smit@tue.nl}
\author[3]{Jianan Zhou*} 
\ead{jianan004@e.ntu.edu.sg}
\author[2,4]{Robbert Reijnen} 
\ead{r.v.j.reijnen@tue.nl}
\author[2,4]{Yaoxin Wu} 
\ead{wyxacc@hotmail.com}
\author[5]{Jian Chen} 
\ead{jchen@nuaa.edu.cn}
\author[3]{Cong Zhang} 
\ead{cong.zhang92@gmail.com}
\author[2,4]{Zaharah Bukhsh} 
\ead{z.bukhsh@tue.nl}
\author[2,4]{Yingqian Zhang}
\ead{YQZhang@tue.nl}
\author[1,2]{Wim Nuijten}
\ead{W.P.M.Nuijten@tue.nl}

\cortext[cor1]{Equal contribution}
\address[1]{Department of Mathematics and Computer Science, Eindhoven University of Technology, Netherlands}
\address[2]{Eindhoven Artificial Intelligence Systems Institute, Eindhoven University of Technology, Netherlands}
\address[3]{College of Computing and Data Science, Nanyang Technological University, Singapore}
\address[4]{Department of Industrial Engineering \& Innovation Sciences, Eindhoven University of Technology, Netherlands}
\address[5]{College of Economics and Management, Nanjing University of Aeronautics and Astronautics, China}

\begin{abstract}
Job shop scheduling problems (JSSPs) represent a critical and challenging class of combinatorial optimization problems. Recent years have witnessed a rapid increase in the application of graph neural networks (GNNs) to solve JSSPs, albeit lacking a systematic survey of the relevant literature. This paper aims to thoroughly review prevailing GNN methods for different types of JSSPs and the closely related flow-shop scheduling problems (FSPs), especially those leveraging deep reinforcement learning (DRL). We begin by presenting the graph representations of various JSSPs, followed by an introduction to the most commonly used GNN architectures. We then review current GNN-based methods for each problem type, highlighting key technical elements such as graph representations, GNN architectures, GNN tasks, and training algorithms. Finally, we summarize and analyze the advantages and limitations of GNNs in solving JSSPs and provide potential future research opportunities. We hope this survey can motivate and inspire innovative approaches for more powerful GNN-based approaches in tackling JSSPs and other scheduling problems.
\end{abstract}




\end{frontmatter}


\newpage
\section{Introduction}
\label{sec:intro}
Job shop scheduling problems (JSSPs) are a significant class of NP-hard scheduling problems.
While being challenging, JSSPs are broadly investigated in computer science and operations research, with various applications in manufacturing~\citep{gupta2006job,zhang2019review,wang2021dynamic}, healthcare~\citep{pham2008surgical,sarfaraj2021applying,lachtar2023application}, and supply chain management~\citep{liu2016parallel,liao2019optimization,cai2023real}. 
Traditional methods for solving JSSPs are mainly exact and heuristic algorithms~\citep{jain1999deterministic,chaudhry2016research,xie2019review,baptiste2001constraint,xiong2022survey}. 
Exact algorithms are designed to find optimal solutions but can suffer from inefficiency and limited scalability. In contrast, heuristic algorithms are more efficient and scalable, yet they may compromise the quality of the found solution.
These two classes of approaches rely heavily on domain knowledge and modeling efforts to identify key parameters, important constraints between parameters, and objective functions. 

In recent years, another class of approach, i.e., machine learning-based methods, has gained increasing attention from researchers. In particular, graph neural networks (GNNs) have been extensively applied to solve JSSPs and have demonstrated promising results. The rationale behind such applications is twofold: 1) Real-world JSSPs need to be solved daily, with massive data of historical instances available. These data can be leveraged for training neural networks, enabling the automatic learning of heuristics for efficiently solving a set of problem instances. 2) JSSPs can be suitably represented by graphs, with the relationships between jobs and machines depicted by the graph topology. GNNs are adept at capturing favorable representations of JSSP instances, which aids in learning effective heuristics for solving them. Despite numerous GNN-based methods being proposed for JSSPs, there is a notable absence of a comprehensive review to gauge the real progress made by these advanced deep learning methods.
This paper aims to fill this gap by reviewing the prevailing GNN-based methods for various types of JSSPs, including the basic JSSP, flexible JSSP, dynamic JSSP, and distributed JSSP, as well as similar problems like the flow-shop scheduling problem (FSP), hybrid FSP, and permutation FSP.

Although several review papers on the application of machine learning (ML) in solving JSSPs exist, a comprehensive review specifically focused on the advanced GNN-based methods for JSSPs is still lacking.
\citet{ccalics2015research} and \citet{zhang2019review} review basic ML-based methods combined with traditional heuristic algorithms for JSSPs, overlooking recent advancements of deep learning methods.
\citet{cunha2020deep} provide a conceptual framework for applying deep reinforcement learning (DRL) to solve JSSPs without delving into further discussions on the neural architectures and training algorithms that are applicable to JSSPs. \citet{li2021machine} present a survey from a bibliometric perspective, offering general statistics on ML-based solutions for JSSPs without a comprehensive overview of the technical details.
More recently, \citet{zhang2023survey} discuss mixed genetic programming and ML techniques for JSSPs, and \citet{zhang2023review} review deep learning techniques for vehicle routing, job shop scheduling, and bin packing problems from a manufacturing perspective. \citet{otala2021graph} summarize the traditional applications of graphs in shop scheduling problems but ignore the power of GNNs to automate policy learning.
These reviews provide limited descriptions for each problem type, often omitting detailed technical insights, particularly on how GNNs are structured and utilized in each method.
Instead, in this paper, we focus on reviewing GNN-based methods for solving different types of JSSPs. The rationale for such a survey is multifaceted. 
Firstly, GNNs have been the most researched models in the realm of deep learning for JSSPs, as instances of JSSPs are characterized by clear graph topologies. 
Secondly, GNN-based methods for JSSPs can facilitate the development of deep learning techniques for other scheduling problems. For example, GNNs with DRL training algorithms for JSSPs can expedite the development of similar methods for machine scheduling problems and task scheduling problems, as highlighted in~\citet{kayhan2023reinforcement,liu2024ga}. Lastly, while surveys on GNN applications in broader combinatorial optimization domains are available~\citep{huang2019review,peng2021graph,cappart2023combinatorial}, there is a lack of review on GNN applications in a specific problem such as JSSPs. 
Given the substantial recent proposals of GNN-based methods for JSSPs, it is critical to provide a fine-grained survey of this rapidly evolving domain. This focused review, offering insights into the development of GNNs for JSSPs, promises to be more valuable and inspiring than a generic list of literature across the entire optimization domain.

In this paper, we explore different GNN models for solving JSSPs and categorize the research based on the training algorithms, namely DRL and non-DRL approaches. This categorization aims to reflect how GNNs can be applied across different training paradigms. Additionally, we extend our survey to include several FSPs that bear similarities to JSSPs, with the goal of inspiring the research community regarding the versatile applications of GNNs in solving different types of scheduling problems and highlighting the disparity between these applications. In summary, the main contributions of this survey paper are outlined as follows:
\begin{itemize}
    \item We describe typical JSSPs from the graph perspective by reviewing the graph representations of different types of JSSPs. Building on this foundation, we introduce the commonly used GNNs for solving JSSPs. This allows us to summarize the relationships and differences between the graph representations of disparate JSSPs and gain insights into how the graph topologies of JSSPs are learned by GNNs.
    
    \item We review existing GNN-based methods for solving various types of JSSPs, categorizing them into groups based on the training paradigm (i.e., DRL and non-DRL). In our descriptions of each method, we explicitly provide critical technical details for GNN applications, such as graph representations, GNN types, GNN tasks, and training algorithms. Additionally, we extend our survey to include similar FSPs, aiming to showcase the methodology of applying GNNs to different scheduling problems and reflecting their broader applicability.
    
    \item After an extensive review of the literature on JSSPs and FSPs, we summarize the advantages and limitations of current GNN-based methods. From this analysis, we propose several promising directions for future research in the application of GNNs to JSSPs. 
\end{itemize}

\section{Problem Description}
\label{sec:pre}
Generally, scheduling problems can be represented by a directed acyclic graph (DAG) consisting of operations (nodes) and precedence constraints (edges), which denote a partial order of the operations to be executed (i.e., the current operation can only start after the precedent operation is finished). The task is to run operations on a set of machines without violating the precedence constraints and others (e.g., resource limit), if any. Each machine can only run one operation at a time. Below, we present a detailed description of each problem type.

\textbf{Open-shop Scheduling Problem.} In open-shop scheduling problem (OSP), all jobs must be processed on a set of machines, with the objective of minimizing the total time to complete all jobs, known as makespan. There are no predefined orders in which jobs must visit the machines, and each job must be processed once on each machine.

\textbf{Flow-shop Scheduling Problem.} Building upon the OSP, the flow-shop scheduling problem (FSP) requires that each job is processed on a set of machines, while the machine sequence is the same and fixed for all jobs. A hybrid flow-shop scheduling problem (HFSP) is a generalization of FSP where each stage can have one or more parallel machines. Jobs must still go through each stage in a fixed sequence, but they can be processed on any machine within each stage. For permutation flow-shop scheduling problem (PFSP), not only must all jobs go through the machines in the same order, but the order of jobs must also be the same on all machines.

\begin{figure}[!t]
    \begin{center}
    \centerline{\includegraphics[width=0.99\columnwidth]{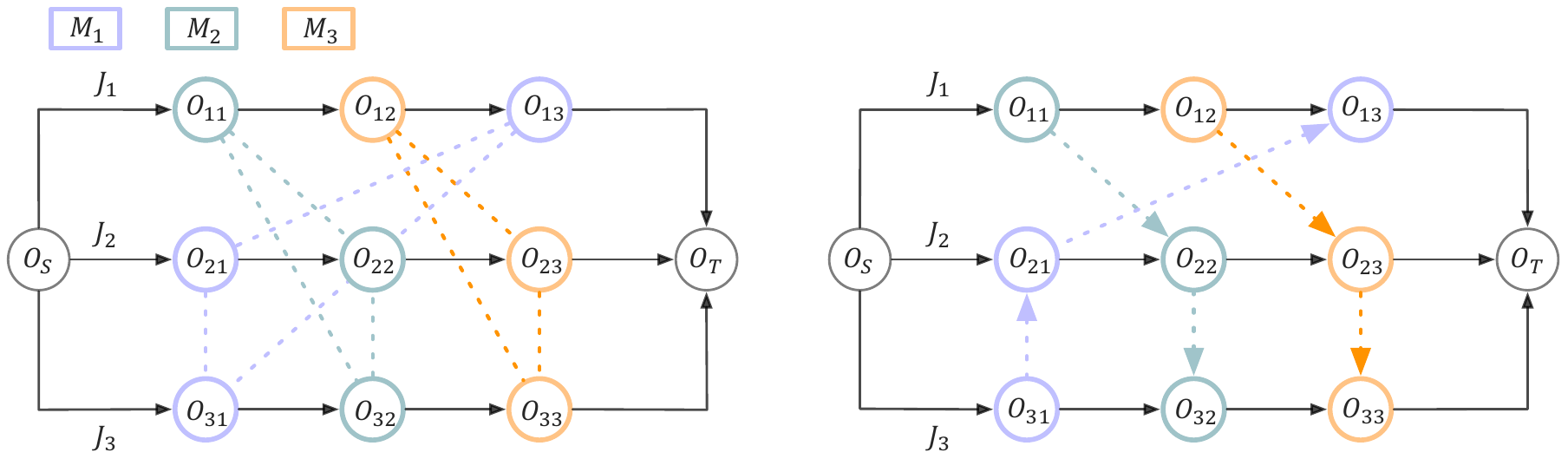}}
    \caption{\textbf{Disjunctive graph representation of JSSPs.} 
    \emph{Left panel} represents a 3 (jobs) $\times$ 3 (machines) JSSP instance. The black arrows are conjunctive arcs, representing the precedence among operations within the same job. The dotted lines are disjunctive arcs whose directions are to be assigned. The disjunctive arcs (or the operation nodes) with the same color require the same machine for processing. 
    \emph{Right panel} represents a feasible solution. Best viewed in color.}
    \label{disjunctive_graph}
    \end{center}
    \vskip -0.3in
\end{figure}

\textbf{Job-shop Scheduling Problem.} Unlike FSPs, each job has a unique sequence of operations that must be processed on specific machines in the job-shop scheduling problem (JSSP). 
Formally, a standard JSSP instance consists of a set of jobs $\mathcal{J}$ and a set of machines $\mathcal{M}$. 
Each job $J_i \in \mathcal{J}$ has an operation set containing $n_i$ operations $O_i=\{O_{ij}\}_{j=1}^{n_i}$ that must be processed in a specific order (i.e. precedence constraints), represented as $O_{i1} \to \dots \to O_{i{n_i}}$.
Here, each $O_{ij}$ signifies an operation of $J_i$ with a processing time $p_{ij} \in \mathbb{N}$.
Each machine can only process one job at a time, and preemption is not allowed. 
The objective in solving a JSSP instance is to determine a start time $S_{ij}$ for each operation $O_{ij}$ such that the makespan $C_{\max} = \max_{i,j} \{C_{ij} = S_{ij} + p_{ij}\}$ is minimized while adhering to all constraints. The size of a JSSP instance is $|\mathcal{J}| \times |\mathcal{M}|$.

For JSSPs, most works employ disjunctive graphs~\citep{NEURIPS2020_11958dfe,park2021learning,zhang2024deep} or augmented graphs with artificial machine nodes~\citep{park2021schedulenet}. Below, we introduce the mainstream disjunctive graph representation. Formally, a JSSP instance is represented by a disjunctive graph $G=(\mathcal{O}, C, D)$~\citep{blazewicz2000disjunctive}. $\mathcal{O}=\{O_{ij}|\forall i,j\} \cup \{O_S, O_T\}$ is the set of all operations, where $O_S$ and $O_T$ are dummy operations with zero processing time, representing the start and terminal states. $C$ is a set of directed arcs (conjunctions) denoting precedence constraints among operations within the same job, while $D$ is a set of undirected arcs (disjunctions) linking operations necessitating the same machine for processing. Consequently, solving a JSSP instance entails determining the direction of each disjunction, ensuring the resultant graph forms a DAG. The longest path from $O_S$ to $O_T$ in a solution is called the critical path, whose length is the makespan of the solution. An example of the disjunctive graph and a feasible solution of a JSSP instance is illustrated in Fig. \ref{disjunctive_graph}. 

\begin{figure}[!t]
    \begin{center}
    \centerline{\includegraphics[width=0.99\columnwidth]{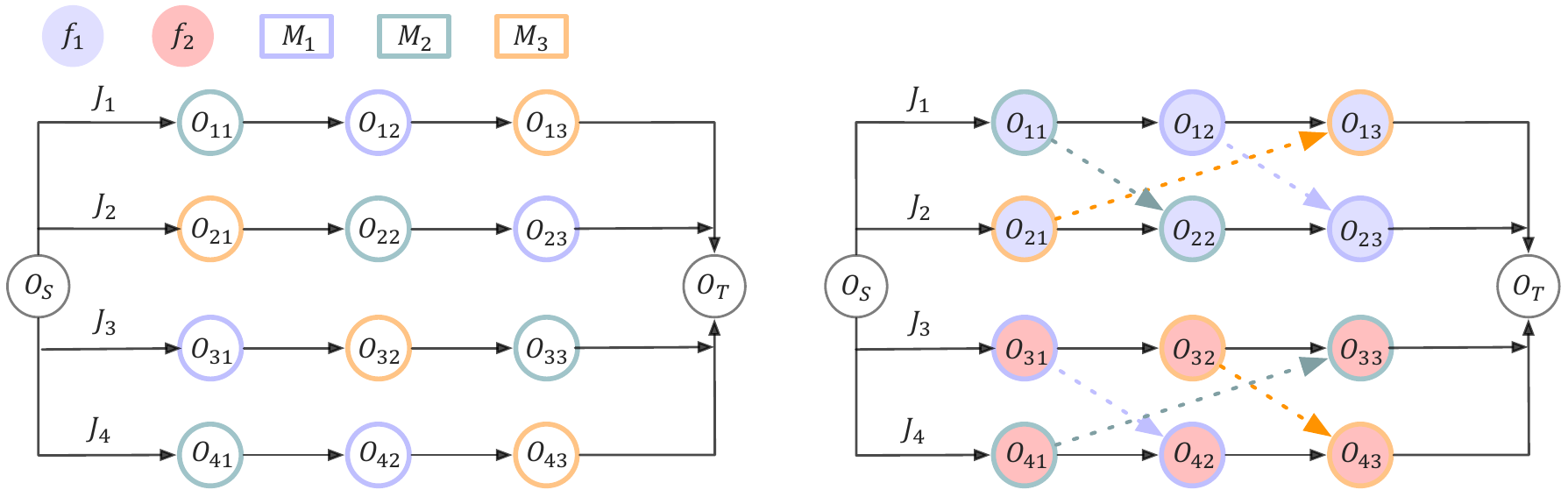}}
    \caption{\textbf{Disjunctive graph representation of Distributed JSSPs.} 
    \emph{Left panel} represents a 4 (jobs) $\times$ 3 (machines) DiJSSP instance. All jobs should be processed in either $f_1$ or $f_2$.
    \emph{Right panel} represents a feasible solution. Best viewed in color.}
    \label{dijssp_disjunctive_graph}
    \end{center}
    \vskip -0.3in
\end{figure}

\textbf{Distributed Job-shop Scheduling Problem.} Distributed job shop scheduling problem (DiJSSP) is a variant of the basic JSSP that incorporates spatial distribution among the machines involved. Formally, a DiJSSP instance comprises a set of factories $\mathcal{F}=\{f_i\}_{i=1}^{|\mathcal{F}|}$, each equipped with its own machines. The problem involves a collection of jobs that must be processed in one factory. 
In DiJSSP, two key decisions should be made: firstly, selecting and assigning each job to a suitable factory, and secondly, scheduling the operations of the jobs within their assigned factory. Once a job begins processing in a selected factory, it cannot be transferred to another. The primary objective is to minimize the maximum makespan across all factories. 
An example of the disjunctive graph representation and a feasible solution of a DiJSSP instance is illustrated in Fig. \ref{dijssp_disjunctive_graph}. 

\begin{figure}[!t]
    \begin{center}
    \centerline{\includegraphics[width=0.99\columnwidth]{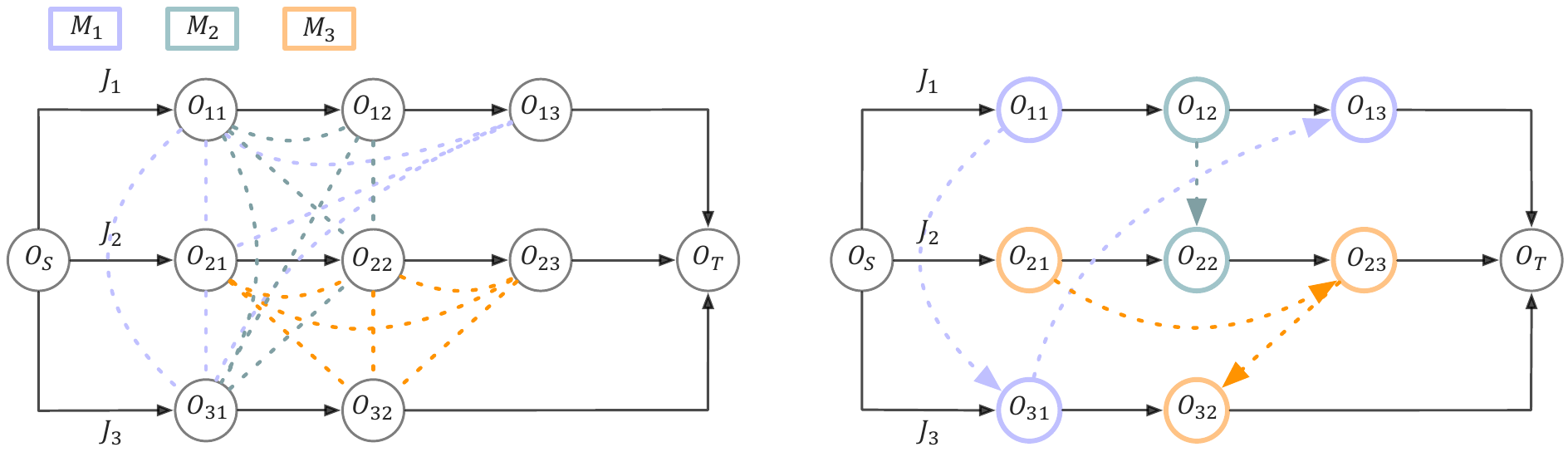}}
    \caption{\textbf{Disjunctive graph representation of Flexible JSSPs.} 
    \emph{Left panel} represents a 3 (jobs) $\times$ 3 (machines) FJSSP instance. Each operation can be processed on any machine from a subset
    of available machines capable of performing that operation.
    \emph{Right panel} represents a feasible solution. Best viewed in color.}
    \label{fjssp_disjunctive_graph}
    \end{center}
    \vskip -0.3in
\end{figure}

\textbf{Flexible Job-shop Scheduling Problem.} 
Different from JSSPs, the flexible job shop scheduling problem (FJSSP) allows each operation to be processed on any machine from a subset of available machines capable of performing that operation.
Formally, each operation $O_{ij}$ can be processed on any machine $M_k$ from its compatible set $M_k \in \mathcal{M}_{ij} \subseteq \mathcal{M}$ with a processing time $p_{ijk}$. Each operation could be connected to multiple disjunctive arcs in the disjunctive graph representation, and thus, solving FJSSP is equivalent to selecting a disjunctive arc for each operation node and fixing its direction. An example of the disjunctive graph and a feasible solution of an FJSSP instance is illustrated in Fig. \ref{fjssp_disjunctive_graph}. 

\textbf{Dynamic Job-shop Scheduling Problem.} Dynamic job shop scheduling problem (DyJSSP) adds a layer of complexity to the traditional JSSP by introducing elements of change and uncertainty that occur over time. Concretely, during the execution of the schedule, new jobs may arrive unpredictably, or uncertain events may happen, such as machine breakdowns, job cancellations, or variations in job processing times. This requires the schedule to be flexible and often necessitates real-time adjustments.

\textbf{Dynamic Flexible Job-shop Scheduling Problem.} By incorporating both flexibility in job routing from FJSSP and dynamic changes in the production environment from DyJSSP, the dynamic flexible job-shop scheduling problem (DyFJSSP) offers a realistic model for optimizing production in complex, dynamic environments. Its ability to respond to real-time changes makes it suitable for modern industries that require high levels of efficiency and adaptability, such as automotive assembly lines, semiconductor manufacturing, and custom product fabrication.

In addition to directed acyclic graphs, there is a recent interest in leveraging alternative graph-based techniques, such as Petri nets, to represent scheduling problems in a more dynamic and concurrent framework. One advantage of Petri nets is their direct interaction with deep reinforcement learning (DRL) agents. Unlike disjunctive graphs, which require additional methods to extract features for optimization, Petri nets naturally integrate with DRL by allowing token distributions to serve as part of the system's observation space. This capability makes Petri nets a promising approach for solving scheduling problems \citep{lassoued2024introducing}.

\section{Graph Neural Networks}
Graph neural networks (GNNs) are a family of deep neural networks that can learn a representation of graph-structured data~\citep{battaglia2018relational}. 
Most GNNs are categorized into four classes. 
1) \emph{Recurrent Graph Neural Network:} Recurrent GNN aims to learn node representations with recurrent neural architectures.
It leverages recurrent connections to capture temporal dependencies within the data and utilizes graph-based operations to handle relationships between entities represented as nodes in the graph.
2) \emph{Convolutional Graph Neural Network:} Convolutional GNN generalizes the operation of convolution from grid data (e.g., images) to graph data. It involves message passing mechanisms to propagate information between nodes, followed by graph convolution operations to update node representations based on their neighborhood structure. This allows it to effectively capture local and global patterns in graph-structured data. Unlike the recurrent GNN iteratively using the same graph recurrent layer in updating node representations, it applies different graph convolutional layers to extract high-level node representations effectively.
3) \emph{Graph Autoencoder:} Graph autoencoder encodes nodes or graphs into a latent representation and reconstructs graph data from the encoded information in an unsupervised learning manner. 
By learning an effective representation in the latent space, it can be used for various tasks such as graph generation and denoising.
4) \emph{Spatial–Temporal Graph Neural Network:} Spatial–Temporal GNN aims to learn hidden patterns from spatial-temporal graphs. Generally, it considers spatial dependence and temporal dependence simultaneously by integrating graph convolutions to capture spatial dependence with RNNs or CNNs to model temporal dependence. They are particularly suited for tasks where spatial relationships and temporal dynamics are crucial, such as traffic prediction, climate modeling, and human activity recognition.
We refer interested readers to~\citet{wu2020comprehensive} for a comprehensive review regarding GNNs. 

Due to the efficiency, generality, and effectiveness of convolutional GNN, its popularity has been rapidly growing in a wide spectrum of applications, including solving NP-hard combinatorial optimization problems. In this survey, we found convolutional GNNs are often used to solve JSSPs. Formally, a graph is represented as $G = \{V, E\}$, where $V$ and $E$ are the sets of nodes and edges, respectively. Let $v_i \in V$ denote a node, and $e_{ij} \in E$ denote an edge pointing from node $v_i$ to $v_j$, where the neighborhood of node $v_i$ is defined as $N(v_i) = \{v_j|e_{ji} \in E\}$. Typically, GNNs use the graph structure and node features to learn a node representation $h_{v_i}$ through the principle of neighborhood aggregation, where they iteratively update each node's representation by aggregating the information of its neighbors. 
After undergoing $k$ iterations of aggregation, the node representation encapsulates the structural information presented in its k-hop network vicinity:
\begin{equation}
    h^{(k)}_{v_i} = \mathcal{C}^{(k)}\left( h_{v_i}^{(k-1)}, \mathcal{A}^{(k)}(\{h_{v_j}^{(k-1)}, \forall v_j\in N(v_i)\}) \right),
\end{equation}
where $\mathcal{A}^{(k)}$ and $\mathcal{C}^{(k)}$ are the aggregation and combination operators at the $k^{\rm{th}}$ layer. They have distinct implementations for different neural networks. 

\textbf{Graph convolutional network (GCN)}~\citep{kipf2017semisupervised} is a powerful architecture motivated by a localized first-order approximation of spectral graph convolutions. The node representation matrix $H$ is updated as follows: 
\begin{equation}
    H^{(k)} = \sigma \left( \tilde{D}^{-\frac{1}{2}} \tilde{A} \tilde{D}^{-\frac{1}{2}} H^{(k-1)} W^{(k)} \right),
\end{equation}
where $\sigma$ is an activation function, $\tilde{A} = A+I$ is the adjacency matrix of the undirected graph with added self-connections, $\tilde{D}_{ii} = \sum_j \tilde{A}_{ij}$, and $W^{(k)}$ is a trainable parameter matrix at the $k^{\rm{th}}$ layer. Unfortunately, it does not naturally support edge features and directed graphs (e.g., disjunctive graphs). 

\textbf{Graph attention network (GAT)}~\citep{veličković2018graph} employs the multi-head attention mechanisms from Transformer~\citep{vaswani2017attention}, with the update formulation as follows:
\begin{equation}
    h_{v_i}^{(k)} = \parallel_{m=1}^M \sigma \left(\sum_{v_j \in N(v_i)} \alpha_{ij}^{m} W^{(k)}_m h_{v_j}^{(k-1)} \right), 
\end{equation}
where $\parallel$ is the concatenation operator, $M$ is the number of attention heads, and $\alpha_{ij}$ is the attention weight that is calculated by:
\begin{equation}
\alpha_{ij} = \frac{\exp \left(\text{LeakyReLU}(W_2^T [W_1h_{v_i} \parallel W_1h_{v_j}]) \right)}{\sum_{v_u \in N(v_i)} \exp \left(\text{LeakyReLU}(W_2^T [W_1h_{v_i} \parallel W_1h_{v_u}]) \right)}
\end{equation}

By leaving out computing $\alpha_{ij}$ if $e_{ji}$
is not present,
GAT can deal with directed graphs. 

\textbf{Graph isomorphism network (GIN)}~\citep{xu2018how} is another popular convolutional GNN variant with strong discriminative power for non-isomorphic graphs. Through theoretical analyses of the expressive power of GNNs, it is demonstrable that GIN stands out as the most expressive among the class of GNNs, matching the power of the Weisfeiler-Lehman graph isomorphism test.
Concretely, it updates the node representation as:
\begin{equation}
    h_{v_i}^{(k)} = \text{MLP}^{(k)}\left( (1+\epsilon^{(k)}) h_{v_i}^{(k-1)} + \sum_{v_j \in N(v_i)} h_{v_j}^{(k-1)} \right),
\end{equation}
where MLP is a Multi-Layer Perceptron, $\epsilon$ is an arbitrary number that can be learned. GIN is empirically found to be effective in solving JSSPs~\citep{NEURIPS2020_11958dfe,zhang2024deep}. 
Most works leveraging GNNs to solve JSSPs directly apply the above-mentioned architecture or design a variant tailored for each specific problem (but following the same principle of neighborhood aggregation), and we refer interested readers to the specific paper.


\section{GNNs for Job Shop Scheduling Problems}
There is a recent trend toward applying GNNs to solve JSSPs. We found that most works apply GNNs with DRL to avoid using optimal solutions as labels. Only a few works employ GNNs without DRL. In this section, we first summarize the GNN-based methods without DRL and then introduce the work on combining DRL and GNN to tackle a variety of JSSPs. At last, we also review the current GNN-based methods for solving flow-shop scheduling problems since they are natural extensions of the methods for JSSPs.

\subsection{Non-DRL Methods}

Several studies have applied GNNs to solve JSSPs without relying on reinforcement learning (RL).
\citet{juros2022exact} evaluate a convolutional GNN-based method, originally proposed by~\citet{gasse2019exact}, on JSSP and the delivery scheduling problem (DSP). The proposed method uses a  convolutional GNN to speed up the variable selection procedure of the branch-and-bound algorithm (B\&B) in the SCIP solver. Their approach employs imitation learning to approximate the full strong branching (FSB) strategy of SCIP for a specific class of problems. This is achieved by solving multiple instances of the same problem type in SCIP and recording the state-action pairs, where actions are the variables chosen for branching by the FSB policy. These pairs are then used to train and test the model, which leverages bipartite representations to model variables and constraints. The results indicate that the model has partially succeeded in learning to imitate the FSB policy. The default branching strategy of SCIP tends to be superior on JSSP, while the learned model solves the problem faster for 13 out of 34 instances on DSP. The authors also report high variance in model performance attributed to the sampling distribution of the problems.

Similar to the previous study, \citet{lee2022imitation} employ imitation learning (i.e., behavior cloning) with a dynamic disjunctive graph representation for the real-time JSSP. The method first generates the dataset by utilizing optimal policies derived from optimal schedules obtained through constraint programming (CP). The dataset captures decision-making points (states) where idle machines have multiple potential (available and soon-to-be-ready) operations (actions) to perform. In this context, the dynamic disjunctive graph adjusts at each decision point, reflecting only the current and imminent operations. Building on this, two GNN models, adapted from~\citet{park2020unsupervised,park2021learning}, are developed. The first GNN utilizes an attention mechanism to aggregate node embeddings from neighbors connected via conjunctive and disjunctive edges. The second GNN, called multiplex, categorizes node features based on three connection types, including input conjunctive, output conjunctive, and disjunctive connections. It processes each category through separate layers and averages these embeddings after several aggregation iterations. The proposed approach is trained on custom-generated instances and tested on TA \citep{TAILLARD_data} benchmark instances for which optimal solutions are known. The results show that the imitation learning approach with disjunctive graphs and multiplex achieves the lowest optimality gap on larger unseen (TA) instances. 



\citet{wang2023graph} explore the use of GNNs and graph transformer models to address combinatorial optimization problems (COPs), specifically focusing on the JSSP and the traveling salesman problem (TSP). Their method employs a two-step learning pipeline: initially, the GNN (or a similar variant) processes a COP instance to generate embeddings, which are then utilized by a functional module to predict the cost or makespan of the optimal solution. For evaluation purposes, benchmark datasets are created for both problems, with the JSSP dataset comprising problem sizes of $9 \times 9$ and $10 \times 10$, containing 10,000 and 100,000 instances respectively, and the TSP dataset including problems with 30 and 40 cities, each having 10,000 instances. The findings indicate that GNN-based models significantly outperform previous models that used convolutional neural networks. However, the paper does not include comparisons with other methods or heuristics. Additionally, the authors suggest leveraging the output from the learning model to guide COP solvers in prioritizing search spaces close to the predicted values, although no experimental results are presented to assess the effectiveness of this strategy.

\citet{Corsini2024} propose a self-supervised learning method, called self-labeling pointer network (SPN) as an alternative to DRL, to create a solution construction agent for JSSPs. In this self-supervised training strategy, multiple parallel solutions are sampled for each problem instance. Then, the best solution is used to provide pseudo-labels, and the cross-entropy loss using these pseudo-labels is minimized. A pointer network using a GAT encoder and a multi-head attention decoder is trained through this method.
In specific, the model is trained using instances generated through Taillard’s method \citep{TAILLARD_data} with sizes of $10\times10$, $15\times10$, $15\times15$, $20\times10$, $20\times15$, and $20\times20$. The method is evaluated on the TA and DMU \citep{DMU_data} benchmark datasets. Their method outperforms priority dispatching rules (PDRs), the insertion algorithm~\citep{Nowicki1996}, learning to dispatch (L2D) ~\citep{NEURIPS2020_11958dfe}, the curriculum learning method~\citep{iklassov2022learning}, and the enhanced local search metaheuristic~\citep{Falkner2022} by a clear margin. Especially when using the sampling inference method, optimality gaps compared to L2D are more than half lower. The performance seems to match, or in some instances outperform, the best DRL-based methods for the JSSP. Thus, this method may offer a promising direction for further research. However, due to the autoregressive inference method without a Markov decision process (MDP) formulation, it may be somewhat limited to static problems. For dynamic problems, a purely autoregressive method may not be suitable without more advanced adaptations. 
Table~\ref{tab:non-drl} provides a summary of GNN-based methods for solving JSSP without using (D)RL. 

\begin{table*}[!t]
\centering
\caption{\textbf{GNN-based non-DRL methods for JSSPs and its variants.}}
\label{tab:non-drl}
\vskip 0.1in
\resizebox{\linewidth}{!}{
\begin{tabular}{@{}c|c|c|c|c|c@{}}
\toprule
\midrule
Method  & Year   & Problem & GNN Task    &   Graph Type    & GNN Type             \\ 
\midrule\midrule
\citet{juros2022exact} & 2022 & JSSP & Variable Selection for Branch and Bound & Bipartite Graph & GCN\\ \midrule
\citet{lee2022imitation} & 2022 & DyJSSP & Dispatching Rule Generation & Disjunctive Graph & Attention and Multiplex \\ \midrule
\citet{wang2023graph}  & 2023 & JSSP & Predict Optimal Cost & Disjunctive Graph & Transformer \\ \midrule
\citet{Corsini2024}  & 2024 & JSSP & Solution Construction & Disjunctive Graph & GAT  \\ \midrule
\bottomrule
\end{tabular}
}
\begin{tablenotes} \scriptsize
		\item The solution construction emphasizes the policy network is not trained to act similarly to a PDR, as it is, for example, trained for solving a single instance or in an autoregressive way.
  
\end{tablenotes}
\label{table:0}
\end{table*}

\subsection{DRL Methods}
By formulating JSSPs as sequential decision-making problems, most existing research on JSSPs leverages GNNs and RL algorithms. On the one hand, the GNN is used to learn a policy to iteratively construct solutions, which is capable of capturing favorable representations of JSSP instances and partial solutions from the topology of graphs. On the other hand, RL algorithms feature interactions between the agent (parameterized by the GNN) and the environment, with trajectories (i.e., sequences of interactions) collected to update the agent. In comparison to supervised learning, RL is beneficial in avoiding the usage of labels, which are optimal solutions and hard to achieve due to the high computational complexity. In the following sections, we will first introduce the preliminaries on DRL, and then elaborate on existing work on applying GNNs and DRL to typical JSSPs. In this paper, we mainly focus on the basic JSSP, flexible JSSP, dynamic JSSP, distributed JSSP, flow-shop scheduling problem (FSP), hybrid flow-shop scheduling problem (HFSP), and permutation flow-shop scheduling problem (PFSP). Since there is little research on other types of JSSPs, we do not cover those problems in this survey.

\subsubsection{Preliminaries on DRL}
DRL combines traditional RL with deep learning techniques to effectively tackle sequential decision-making problems. Given an MDP formulation of a sequential decision-making problem, an agent sequentially takes actions according to the varying states derived from the environment. The environment iteratively responds to the agent’s actions by providing rewards and transitioning to new states. 
The objective of DRL is to find an optimal policy $\pi^*_{\theta}$ that maximized the expected cumulative reward (i.e., return). Generally, policy-based and value-based approaches are two primary methods for learning optimal policies. 
Policy-based approaches, such as REINFORCE, A3C, and PPO~\citep{schulman2017proximal,williams1992simple,mnih2016asynchronous}, directly learns the policy $\pi_{\theta}$, which defines the probability distribution over actions for each state. 
In contrast, value-based approaches, such as DQN, double DQN, and dueling DQN~\citep{wang2016dueling,mnih2015human,van2016deep}, focuses on learning value functions $Q^*_{\delta}$, which estimate the expected cumulative reward for each state (or state-action pair).
We refer interested readers to \citet{franccois2018introduction,tassel2021reinforcement} for more knowledge on DRL.




\subsubsection{MDP formulations of JSSPs}
The MDP formulations for JSSPs are varying in different methods. Here, we introduce one commonly used MDP formulation for learning construction heuristics in most existing literature~\citep{NEURIPS2020_11958dfe}. In the MDP, 1) the \emph{state} is the problem instance and the partial schedule updated in the solving process, which is depicted by the graph representation introduced in Section~\ref{sec:pre}; 2) given the current state, the \emph{action} set is dispatching operations to their designated machines; 3) the \emph{transition} is a deterministic function, which changes the state (i.e., the graph representation) according to the dispatching action (e.g., a new link can be added in the graph to represent an operation is dispatched after another operation on the same machine); 4) the \emph{reward} can be defined as the difference between the makespan of the current state and the previous state, with the return of a trajectory equivalent to the total makespan (i.e., the objective); 5) the stochastic \emph{policy} takes as input the graph representation of state, and outputs feasible operations to be dispatched to a machine. On top of the above general MDP, various MDPs are also proposed for different types of JSSPs with slight adjustments. For example, the commonly used MDP of FJSSPs only differs slightly from JSSPs in the state and action set. Specifically, the state representation includes additional machine nodes, often depicted as extra nodes in a heterogeneous graph~\citep{song2022flexible}.
Actions, in turn, consist of a selected operation-machine pair, rather than merely an operation, after which the corresponding links are adjusted in the transition function.

In DRL-based methods, GNNs play important roles in learning policies for constructing solutions. Various GNNs have been proposed for solving JSSPs with DRL. In the following sections, we will elaborate on how these GNNs are used in each study concerning JSSPs, with details summarized in Table~\ref{table:1}.

\begin{table*}[!t]
\centering
\caption{\textbf{GNN-based DRL methods for basic JSSPs.}}
\vskip 0.1in
\resizebox{\linewidth}{!}{
\begin{tabular}{@{}c|c|c|c|c|c|c@{}}
\toprule
\midrule
Method                             & Year   & Problem & GNN Task    & DRL Algorithm     & Graph Type    & GNN Type            \\ 
\midrule\midrule
\citet{NEURIPS2020_11958dfe}                                & 2020 & JSSP & Dispatching Rule Generation & PPO & Disjunctive Graph & GIN \\ \midrule
\citet{park2021schedulenet}                        & 2021   & JSSP    &  Dispatching Rule Generation & REINFORCE                     & Disjunctive Graph & Type-Aware Graph Attention                \\ \midrule
\citet{park2021learning}                              & 2021   & JSSP    &  Dispatching Rule Generation & PPO                           & Disjunctive Graph     & Message Passing           \\ \midrule
\citet{tassel2021reinforcement}                             & 2021  & JSSP    &  Solution Construction & PPO                           & Gantt Chart & MLP                      \\ \midrule
\citet{hottung2021efficient}                                & 2022 & JSSP    & Dispatching Rule Generation & PPO                           & Disjunctive Graph & GIN               \\ \midrule
\citet{Liu2022} & 2022 & JSSP &Dispatching Rule Generation & Dueling-DDQN & Machine-Job Graph & Custom Graph Embedding \\ \midrule
\citet{Yang2022} & 2022 & JSSP &Dispatching Rule Generation & PPO & Disjunctive Graph & GAT \\ \midrule
\citet{Falkner2022} & 2022 & JSSP &Guiding Local Search & DDQN & Directed Acyclic Graph & GCN \\ \midrule
\citet{Chalumeau2023} & 2023 & JSSP & Dispatching Rule Generation & PPO & Disjuntive Graph & Transformer \\ \midrule
\citet{Liao2022} & 2023 & JSSP & Dispatching Rule Generation & Option-Critic & Disjunctive Graph & Message Passing \\ \midrule
\citet{Hameed2023} & 2023 & JSSP &Dispatching Rule Generation & PPO & Bipartite Graph & Message Passing \\ \midrule
\citet{Ho2023} & 2023 & JSSP & Selecting PDRs & DDQN & Heterogeneous Graph & Heterogeneous GIN \\ \midrule
\citet{Fang2023} & 2023 & JSSP & Selecting PDRs & PPO & Disjunctive Graph & AttentionWalk \\ \midrule
\citet{Shi2023} & 2023 & JSSP & Dispatching Rule Generation & PPO & Disjunctive Graph & GIN \& LSTM\\ \midrule
\citet{Chen2023} & 2023 & JSSP & Dispatching Rule Generation & REINFORCE & Disjunctive Graph & Transformer \\ \midrule
\citet{Liao2023} & 2023 & JSSP & Dispatching Rule Generation & PPO & Matrix & Transformer \\ \midrule
\citet{Chen2023multi} & 2023 & JSSP & Dispatching Rule Generation & PPO & Disjunctive Graph & GIN \\ \midrule
\citet{Yuan2023} & 2023 & JSSP & Dispatching Rule Generation & PPO & Disjunctive Graph & GIN \\ \midrule
\citet{Tassel2023} & 2023 & JSSP & Dispatching Rule Generation & Imitation Learning and Policy Gradient Hybrid & Matrix & Transformer \\ \midrule
\citet{Ho2024} & 2024 & JSSP & Dispatching Rule Generation & REINFORCE & Heterogeneous Graph & Heterogeneous GIN \\ \midrule
\citet{Wong2024} & 2024 & JSSP & Dispatching Rule Generation & PPO & Disjunctive Graph & Message Passing \\ \midrule
\citet{Lee2024} & 2024 & JSSP & Dispatching Rule Generation & REINFORCE & Matrix & Transformer \\ \midrule
\citet{zhang2024deep} & 2024 & JSSP & Learning Improvement Heuristic & REINFORCE & Disjunctive Graph & GIN \& GAT \\ \midrule
\citet{Remmerden2024offline} & 2024 & JSSP & Dispatching Rule Generation & Offline RL with Conservative Qlearning (CQL) & Disjunctive Graph & GIN  \\ \midrule
\bottomrule
\end{tabular}}
\label{table:1}
\end{table*}

\subsubsection{Basic JSSPs}


\citet{NEURIPS2020_11958dfe} are the first to propose a DRL approach, called L2D, to solve JSSPs. They propose to learn priority dispatching rules based on the disjunctive graph representation. They use a GIN network trained with PPO to learn embeddings for the disjunctive graphs and select operations to schedule. The proposed method is trained and evaluated on instances generated by Taillard’s method with sizes ranging between $6\times6$ and $30\times20$. In addition, they report their evaluation results on the TA and DMU benchmark instances up to $100\times20$. Their method outperforms priority dispatching rules while maintaining a fast speed. The method underperforms compared to exact solutions obtained through the OR-Tools CP-SAT solver \citep{cpsatlp}, but it operates much faster for larger problems.
\citet{park2021learning} propose GNNRL, a similar method to L2D. Instead of a GIN network, they design a custom graph embedding layer that incorporates a basic message passing structure, which accounts for different edge types. For training, they generate instances following the uniform distribution of 5 to 9 machines and up to 9 jobs with uniform processing times between 1 to 99 time units. They evaluate with similar instances, as well as the ORB \citep{ORB_data}, SWV \citep{SWV_data}, LA \citep{Lawrence_Data}, ABZ \citep{ABZ_data}, and YN \citep{YN_data} benchmarks, which have sizes from 5 to 20 machines and 6 to 100 jobs. Their method shows better performance than all PDRs and good generalization to the different benchmarks with varying sizes.
\par
\citet{hottung2021efficient} propose an efficient active search (EAS) to improve deep learning policies for combinatorial optimization. This method adjusts a subset of model parameters on an individual instance in order to search for better solutions. They test their method on several combinatorial optimization problems, including JSSPs. In specific, they use EAS to enhance the performance of L2D policies. Their method considerably improves the performance over naive sampling methods at the cost of increased runtime. As EAS is generally applicable, it can be used to improve any of the discussed DRL methods in this paper. \citet{Chalumeau2023} also propose an alternative method to improve existing policies in combinatorial optimization based on latent space search. This method, called COMPASS, conditions the networks on a latent space. Then, the latent space is searched using evolutionary strategies to obtain better policies. Although additional training is needed for COMPASS, the inference time barely increases, unlike EAS. In addition, COMPASS achieves better performance than EAS. Hence, it provides another alternative to improve existing policies. 
\par
In contrast to the previous methods, \citet{park2021schedulenet} consider the JSSP as a multi-agent problem, with machines representing agents. They propose ScheduleNet, which is a general scheduler for solving multi-agent scheduling problems. This approach utilizes an agent-task graph to model relationships and employs a type-aware graph attention network to extract information. This method facilitates efficient scaling to larger settings. They uniformly sample instances for training with 7 to 14 jobs and 2 to 5 machines. The method is evaluated on randomly generated instances ranging from size $6\times6$ to $100\times20$, as well as the ORB, SWV, FT, LA, YN, and TA benchmarks. ScheduleNet outperforms the previous methods~\citep{NEURIPS2020_11958dfe,park2021learning}, while being a generalizable method. In addition, for the $100\times20$ instances, they outperform CP-SAT with a 1-hour time limit.
\par
\citet{tassel2021reinforcement} provide a different approach to utilizing DRL for JSSPs. Their method, JSSEnv, treats each scheduling instance as a unique learning problem that has to be solved from scratch. They allow 10 minutes of learning per problem instance, and select the best solution constructed within this timeframe. They use a compact matrix state representation and define a reward function based on the idle time of the machines. Moreover, they reduce the action space by applying basic scheduling principles. They evaluate their method on TA instances 41-50 and DMU instances 16-20, all featuring 30 jobs and 20 machines. They outperform multiple PDRs and L2D. However, this is an unfair comparison as L2D proposed a generalizable solution that is not fully trained for individual instances. They do not reach the performance of the CP-SAT solver, which is on average 6\% to 7\% better.
\par
\citet{Liu2022} introduce a method employing a dueling double deep Q-network (dueling-DDQN), called G3DQN, to tackle JSSPs. They formulate JSSP instances as a machine-job graph, represented by three matrices: disjunctive, weight, and state matrices. They also propose a custom graph embedding layer. They train their model on random instances of size $5\times5$, $6\times6$, and $10\times5$ with processing times uniformly sampled between 20 and 70. They assess their method on independent instances of the same types and consistently outperform PDRs across each instance size while maintaining high speed. 
For larger instances of up to 25 jobs and 15 machines, they use the DRL solution as an initial solution for a simulated annealing (SA) strategy, demonstrating an improvement in SA performance compared to regular initialization. Additionally, they discuss how their method can be adapted for FJSSPs, although they do not provide empirical evaluations.
\par
\citet{Yang2022} employs the disjunctive graph representation for JSSPs and introduces a method using a GAT combined with PPO learning to construct scheduling solutions. They train and evaluate their method on random instances generated using Taillard’s method with sizes of $15\times15$, $20\times15$, and $20\times20$. Their approach slightly improves over a baseline, which uses a GIN with PPO. They also assess the generalization and find that GIN performs better on $50\times15$, whereas GAT maintains superior results on $50\times20$. While they do not compare their method with other solution strategies, their reported results indicate a slight advantage over the L2D. 
\citet{Liao2022} also use the disjunctive graph representation and explore hierarchical reinforcement learning as a promising strategy to handle scheduling problems with varying instance characteristics. They apply the option-critic algorithm and a message passing neural network over the disjunctive graph. They train and evaluate their method on random instances using Taillard’s methods with sizes $6\times6$, $10\times10$, $15\times15$, and $20\times20$. In addition, they select a handful of instances from the OR-library \citep{OR-library} for further evaluation. Their method outperforms the least operation remaining (LOR) dispatching rule, demonstrating the promise for further advancements of hierarchical RL strategies, possibly through the implementation of better network architectures.
\par
\citet{Hameed2023} adopt a different graph representation for JSSPs, using a bipartite graph consisting of machine nodes and machine buffer nodes. They develop a basic message passing network to train a PPO agent that assigns jobs to the machines. Utilizing PPO along with curriculum learning, they train their agent on a single scheduling instance and evaluate it on that instance, as well as two slight variations of the problem. Their method surpasses both Tabu search and a genetic algorithm on this instance. However, they do not extend the evaluation to other problem instances, which significantly restricts the practical applicability of their method.
\par
\citet{Ho2023} consider a different action space. At each time step, the DRL agent selects one PDR amongst a set of possible PDRs to use for the new dispatching step. They adapt the method originally proposed by \citet{luo2020dynamic} by modifying the state space to a graph format and employing a heterogeneous GIN architecture. This adjustment enhances the generalization capabilities compared to the previous approach. They use the TA instances to verify their performance. Concretely, they outperform all priority dispatching rules and L2D across both small and large instances. Oppositely, they are slightly worse than ScheduleNet on small instances while being marginally better on large instances.
\citet{Fang2023} also consider selecting PDRs to use at each step, applying their method to handle both the JSSP and the JSSP with scheduling maintenance times (JSSP-MT). They use the AttentionWalk method~\citep{abu-el-haija2018} to learn embeddings for the disjunctive graph of a problem instance, independently of the DRL framework. 
The DRL approach then incorporates these attention embeddings along with a set of dynamic features to learn a policy. 
They evaluate their method on several TA instances, which are modified to include maintenance times. Their results show improvements over PDRs and L2D, and also exhibit a larger margin of improvement compared to CP-SAT solutions for both JSSP and JSSP-MT. In a new paper, \citet{Ho2024} significantly expand their previous work~\citep{Ho2023} by applying their heterogeneous GIN architecture in a more elaborate and sophisticated approach, termed residual scheduling. In this approach, they adapt the state space at each step by removing irrelevant information about past activities. Using the REINFORCE algorithm, the DRL agent learns to select machine-node pairs at each time step, making it suitable for JSSPs and FJSSPs. They train the model on instances generated by Taillard’s method with up to 10 machines and 10 jobs. For JSSPs, they utilize the TA, ABZ, FT, ORB, YN, SWV, and LA evaluation benchmarks, covering sizes from $6\times6$ up to $100\times20$, as well as synthetic instances with sizes of $150\times15$ and $200\times15$. For FJSSPs, the Brandimarte \citep{Brandimarte1993} and Hurink \citep{Hurink1994} benchmarks are used. The method generally outperforms L2D, GNNRL, and ScheduleNet on all JSSP instances, and also surpasses the approach by \citet{song2022flexible} on FJSSP instances, with small gaps to the CP-SAT solutions given half a day of computation time. For many JSSP instances with over 100 jobs, their method matches the CP-SAT performance.
\par
\citet{Shi2023} attempt to solve the JSSP using a hybrid network architecture combining GIN with LSTM. However, they do not provide a clear explanation of how these architectures are integrated, which significantly limits the clarity and reproducibility of their method. Similar to many other studies, they use the disjunctive graph representation and PPO learning algorithm. They also train and test their method using random instances generated via Taillard’s method. They outperform both PDRs and L2D across all instance sizes, ranging from $6\times6$ to $100\times20$.
\par
\citet{Chen2023} propose DGERD, a transformer-based construction policy for JSSPs. In their method, they first train a transformer encoder network offline using a strategy akin to node2vec. Then, they employ a REINFORCE-like algorithm to learn a policy that integrates the graph embeddings with additional features at each time step. Their method surpasses traditional PDRs across a range of TA instances. However, they do not provide comparisons of their method with other learning-based approaches.
\citet{Liao2023} also propose a transformer network approach.  Different from \citet{Chen2023}, they train their full network architecture using PPO without offline learning. They train their model on random instances generated through Taillard’s method with sizes of $10\times5$, $10\times10$, $20\times10$, and $20\times20$, and evaluate on the TA benchmark dataset, covering sizes from $15\times15$ up to $100\times20$. They outperform PDRs and L2D on these instances.
\par
\citet{Chen2023multi} develop construction policies for a multi-objective scheduling problem that considers makespan, finish rate, and cost. They apply a GIN network trained with PPO. To aid the training, they employ a method based on the convex hull to adjust the weights of different rewards within the environment. They train their policies on instances defined by \citet{NEURIPS2020_11958dfe}, ranging from $10\times10$ to $30\times20$, where values for the two non-makespan objectives are generated. Their policy achieves better costs and finish rates with only a slight sacrifice in makespan compared to L2D. They also demonstrate that their convex hull method considerably outperforms a straightforward application of DRL on a weighted-sum objective. In addition, they maintain a better makespan over PDRs on both the generated instances and several DMU benchmark instances with sizes of $30\times20$, $50\times15$, and $50\times20$.
Similar to \citet{Chen2023multi}, \citet{Yuan2023} utilize a disjunctive graph representation, GIN, and PPO to develop their solutions. They, however, focus solely on makespan optimization, which aligns with the majority of similar studies. They also integrate the action masking approach suggested by \citet{tassel2021reinforcement} to improve their solution construction. They train various policies for job-shop configurations ranging from 10 to 30 jobs and 10 to 20 machines, with uniform processing times between 1 and 99. They conduct evaluations across several benchmark datasets, including ABZ, FT, ORB, YN, SWV, LA, and TA. Their method demonstrates superior performance compared to GNNRL on most of these datasets, as well as L2D, ~\citet{Chen2023multi}, and several PDRs.
\par
\citet{Tassel2023} propose a hybrid learning approach to train a DRL agent by leveraging existing CP solvers. This method combines imitation learning with policy gradient learning. Concretely, policy gradient training is used to generate good initial solutions, which form the first part of a schedule. Imitation learning is then used to mimic the solutions of CP solvers, aiming to complete these initial solutions. To enhance performance, multiple actors are trained in parallel, and a simulated annealing-like method is used to select operations from these actors. Training is conducted on four TA instances with 30 jobs and 15 machines, which strikes a balance between size and computational cost. For evaluation, they use a range of benchmarks, including TA, DMU, LA, ORB, SWV, YN, and the novel Da Col and Teppan benchmarks, extending up to 1000 jobs and 1000 machines. Their method generally outperforms PDRs and is notably fast. In addition, they compare their results with the Choco CP solver, outperforming it given the same amount of running time, although comparing them within the same time constraints may not be entirely fair due to the differing nature of CP solvers. The empirical results demonstrate good generalizability across various datasets despite the method being trained on only four instances, suggesting this hybrid approach is a promising avenue for future research.
\par
\citet{Wong2024} tackle a complex variant of the JSSP, known as the interrupting swap-allowed blocking JSSP. They model interruptions by dynamically adding and deleting nodes and edges within the disjunctive graph formulation. A dispatching agent is trained with PPO, using a message passing network with heterogeneous layers tailored to different types of node connections. The training involves randomly generated instances with up to 9 machines and 9 jobs. The method is tested on 18 benchmark instances of size $10\times10$, as described by \citet{Mascis2002}. The proposed approach outperforms a variety of PDRs across most instances, effectively handling various rates of interruptions.
\par
\citet{Lee2024} recently propose ARLS, effectively applying a Transformer-like architecture for a DRL-based scheduling algorithm.
They incorporate masked attention to focus on operations within the same job or machine and train their agent using the REINFORCE algorithm on synthetic instances with the size of $6\times6$, similar to those used by \citet{NEURIPS2020_11958dfe}.
The evaluation is conducted on the TA, ABZ, FT, ORB, YN, SWV, and DMU benchmark datasets, ranging from sizes $6\times6$ to $100\times20$. 
Remarkably, despite being trained on such small instances, their method outperforms traditional dispatching rules, L2D, GNNRL, ScheduleNet, DGERD, and \citet{Yuan2023} on most benchmarks by a considerable margin, except a couple of instances at which ScheduleNet is slightly better.
\par
While most research primarily focuses on constructive methods, \citet{Falkner2022} and \citet{zhang2024deep} aim to improve search methods. \citet{Falkner2022} propose NeuroLS, a DRL approach designed to steer local search for combinatorial optimization. They integrate a GNN to capture information on current solutions found by the local search engine, combining it with state information from the local search itself. They train their model using a DDQN approach on instances generated by Taillard’s method, ranging in sizes from $15\times15$ to $30\times20$. The evaluation is conducted on the TA benchmark dataset. With 100 iterations of the local search, NeuroLS outperforms L2D, GNNRL, and ScheduleNet, as well as traditional optimization methods like simulated annealing, iterated local search, and variable neighborhood search (VNS) on most problem sizes, though the improvement over VNS is limited.
\citet{zhang2024deep} opt to use a GNN-based DRL approach to create an improvement heuristic. In their method, the agent guides a neighborhood-based improvement heuristic by selecting node-pairs to swap within the disjunctive graph. They train their method on synthetic instances generated using Taillard’s method, with sizes of $10\times10$, $15\times10$, $15\times15$, $20\times10$, and $20\times15$. The evaluation encompasses the TA, ABZ, FT, LA, SWV, ORB, and YN benchmarks, along with three very large datasets containing up to 1000 jobs. Although this method increases runtime, it significantly outperforms construction-based DRL methods on the benchmarks, achieving near-optimal solutions for some instances. Furthermore, the method proves competitive against a tabu search approach while being faster. Notably, on large datasets, within a few minutes of computation, the method achieves 15 to 25 percent improvements over solutions obtained using CP-SAT with a 1-hour time limit. This method outlines a promising new research direction for the application of DRL to improvement heuristics in scheduling.
\par
Most existing DRL approaches for JSSPs are online RL, where an RL agent selects actions by interacting with an environment.  
One downside of online RL is that it cannot learn from existing data, for instance, solutions generated from optimization algorithms, although such data is easily obtained as many fast heuristics exist. Imitation learning utilizes existing data but is hard to obtain decisions better than the algorithm used to generate training data.   \citet{Remmerden2024offline}  propose the first Offline RL method to solve JSSP,   Offline Reinforcement Learning
for Learning to Dispatch (Offline-LD), which makes use of existing data and overcomes the limitation of imitation learning. Offline-LD is developed based on two CQL-based Q-learning methods (mQRDQN and discrete mSAC), and 
utilizes Conservative Qlearning
(CQL) to enable d-mSAC and 
mQRDQN to learn from existing datasets. They show that with Offline-LD, the existing online RL approach, such as L2D  \cite{NEURIPS2020_11958dfe}, can be easily adapted as an offline RL method. 
The experiments demonstrate  
that Offline-LD, trained with only 100 instances, outperforms online RL (L2D), behavioral cloning approach, and dispatching rules on generated and benchmark instances. In addition, the results show that less perfect data, generated with noises, can achieve similar or better policies than those obtained from  
the expert dataset generated by a CP solver.  This work provides a new promising RL approach to solving JSSP and its variants.

\begin{table*}[!t]
\centering
\caption{\textbf{GNN-based DRL methods for FJSSPs.}}
\vskip 0.1in
\resizebox{\linewidth}{!}{
\begin{tabular}{@{}c|c|c|c|c|c|c@{}}
\toprule
\midrule
Method                             & Year   & Problem & GNN Task    & DRL Algorithm     & Graph Type    & GNN Type             \\ 
\midrule\midrule

\citet{lei2022multi} & 2022 & FJSSP & Dispatching Rule Generation & Multi-PPO & Disjunctive Graph & GIN \\ \midrule
\citet{zeng2022deep} & 2022 & FJSSP & Dispatching Rule Generation & A3C & Disjunctive Graph & GIN \\ \midrule
\citet{song2022flexible} & 2022 & FJSSP & Dispatching Rule Generation & PPO & Heterogeneous Graph & Heterogeneous GAT \\ \midrule
\citet{dax2022graph} & 2022 & FJSSP & Dispatching Rule Generation & Q-learning & Heterogeneous Graph & Heterogeneous GCN \\ \midrule
\citet{zhang2023reinforcement} & 2023 & FJSSP & Dispatching Rule Generation & Actor-Critic & Heterogeneous Graph & Hierarchical GNN \\ \midrule
\citet{farahani2023relational} & 2023 & FJSSP & Dispatching Rule Generation & Advantage Actor Critic & Disjunctive Graph & GAN \\ \midrule
\citet{wang2023flexible} & 2023 & FJSSP & Dispatching Rule Generation & PPO & Disjunctive Graph & GAT \\ \midrule
\citet{oh2023applying} & 2023 & FJSSP & Dispatching Rule Generation & DQN & Machine Pairwise Graph & GAT \\ \midrule
\citet{zhang2023deepmag}  & 2023 & FJSSP & Dispatching Rule Generation & DQN & Multi-Agent Graph & MLP \\ \midrule
\citet{jing2024multi} & 2024 & FJSSP & Dispatching Rule Generation & Multi-agent Reinforce & Directed Acyclic Graph & GCN \\
\midrule
\bottomrule
\end{tabular}}
\label{table:2}
\end{table*}

\subsubsection{Flexible JSSPs}
\par In consideration of system complexities, GNNs have increasingly been employed to synergize feature extraction with reinforcement learning techniques. 
\cite{zeng2022deep} formulate the FJSSP as an MDP where the disjunctive graph is used to represent the state, and operation set and machine allocation are used as the actions. A powerful GIN is used to extract the state features. The asynchronous advantage actor-critic (A3C) algorithm is selected for optimization training, with performance evidenced on Brandimarte instances.
\cite{dax2022graph} use Q-Learning to train GNNs for tackling the FJSSP. They utilize disjunctive graphs to simulate states and encode an action value function with heterogeneous GCNs. Benchmark testing shows the algorithm's advantages in terms of training time and solution quality compared to metaheuristic algorithms.
\cite{song2022flexible} propose an end-to-end DRL algorithm to learn PDRs for the FJSSP. An MDP formulation integrates the operation selection and machine assignment into a single decision. They expand the disjunctive graph with machine nodes and introduce a heterogeneous graph to describe the state. The PPO algorithm is applied to train the policy network. Experiments show their approach outperforms traditional PDRs.
\cite{zhang2023reinforcement} propose a hierarchical graph structure model that synergistically integrates the environment and decision-making module. They formulate the environment module as a heterogeneous graph to define multi-relational FJSSP instances and use a hierarchical GNN. The decision module incorporates actor-critic networks to train reasoning strategies and enhance decision performance. This approach outperforms PDRs and DRL-based methods in Brandimarte and Hurink test cases.
\cite{farahani2023relational} propose an edge feature guided relational graph attention-based deep reinforcement learning (ERGAT-DRL) framework to address the FJSSP considering the sequence of setup times. A weighted relational graph is employed to address the machine flexibility and sequence dependency. They utilize an advantage actor-critic (A2C) algorithm to train their model. Their method outperforms PDRs on realistic data.
\cite{wang2023flexible} formulate the FJSSP using a tight state representation for describing operations and machines. Additionally, an end-to-end learning framework, which combines self-attention models with DRL, is introduced. A dual attention network (DAN) for extracting features between operations and machines is trained using PPO in the training phase. Experiments on synthetic and public benchmark datasets from Brandimarte and Hurink have confirmed the superiority over traditional PDRs and state-of-the-art DRL methods.
\cite{lei2022multi} introduce a multi-pointer graph network (MPGN) architecture to embed local states effectively. This architecture delineates two sub-policies: job operation action and machine action. During the training phase, a multi-PPO approach is adopted to learn these sub-policies. Benchmark results show that the proposed method achieves superior generalization performance. 
\par
Single-agent reinforcement learning (SARL) simplifies computational demands but encounters difficulties with complex and dynamic scheduling tasks. Another research stream emphasizes the application of GNNs in multi-agent reinforcement learning (MARL) to tackle the FJSSP. The integration of MARL with GNNs aims to enhance scheduling performance and boost system autonomy and scalability through the exploitation of intricate job-machine interactions. However, this integration faces challenges such as high computational loads and the complexity of distilling essential insights from vast datasets. 
\cite{jing2024multi} model the FJSSP as a topoligical graph structure prediction process and develop a GCN-based multi-agent system. A centralized-learning decentralized-execution scheduling structure is proposed to balance the contradiction between the autonomy and stability of a multi-agent system. The experimental results demonstrate algorithmic superiority on the Brandimarte and Hurink datasets. 
\cite{oh2023applying} introduce a multi-agent system for the FJSSP with a dynamic number of jobs. To address the scalability issue, they model FJSSP as a machine pairwise graph structure and use GNNs to reflect cooperative agent relationships. The distributional value decomposition network (DDN) algorithm is used for training, ensuring utility maximization for each agent in the multi-agent network.
\cite{zhang2023deepmag} formulate a new model for the FJSSP by combining DRL and MARL. They design a multi-agent graph to derive the relationships among machines and jobs. The machines and jobs are regarded as agents and represented by nodes, with edges reflecting the order of machines processing operations and the operations that can be processed.  Moreover, DQN and MARL are used to manage state and action spaces effectively. They demonstrate their superior performance in real-world instances compared to other state-of-the-art methods.

\begin{table*}[!t]
\centering
\caption{\textbf{GNN-based DRL methods for DyJSSPs and DiJSSPs.}}
\vskip 0.1in
\resizebox{\linewidth}{!}{
\begin{tabular}{@{}c|c|c|c|c|c|c@{}}
\toprule
\midrule
Method                             & Year   & Problem & GNN Task    & DRL Algorithm     & Graph Type    & GNN Type             \\ 
\midrule\midrule 
\citet{zeng2022hybrid} & 2022 & DyJSSP & Selecting PDRs & D3QPN & Disjunctive Graph & GAT \\ \midrule
\citet{liu2023dynamic} & 2023 & DyJSSP & Dispatching Rule Generation & PPO & Disjunctive Graph & Message Passing \\ \midrule
\citet{yang2023combining} & 2023 & DyJSSP & Selecting PDRs & DDQN & Disjunctive Graph & GCN \\ \midrule
\citet{zeng2023you} & 2023 & DyJSSP & Selecting PDRs & D3QPN & Disjunctive Graph & Transformer \\ \midrule
\citet{su2023evolution} & 2023 & DyJSSP & Dispatching Rule Generation & Evolution Strategies & Disjunctive Graph & GIN \\ \midrule
\citet{zhang2023deep} & 2023 & DyFJSSP & Dispatching Rule Generation & PPO & Disjunctive Graph & GIN \\ \midrule
\citet{lei2023large} & 2023 & DyFJSSP & Dispatching Rule Generation & Multi-PPO & Disjunctive Graph & GIN \\ \midrule 
\citet{zhang2023deep} & 2023 & DyFJSSP & Dispatching Rule Generation & PPO & Disjunctive Graph & MLP \\ \midrule
\citet{zhang2023dynamic} & 2023 & DyFJSSP & Dispatching Rule Generation & PPO & Heterogeneous Graph & Heterogeneous GAT \\ \midrule
\citet{huang2023novel} & 2023 & DiJSSP & Dispatching Rule Generation & PPO & Disjunctive Graph & GIN \\ \midrule
\citet{huang2024end} & 2024 & DiJSSP & Dispatching Rule Generation & PPO & Disjunctive Graph & GIN \\ \midrule
\citet{huang2024hierarchical} & 2024 & DiJSSP & Dispatching Rule Generation & PPO & Disjunctive Graph & GIN \\ 
\midrule
\bottomrule
\end{tabular}}
\label{table:3}
\end{table*}

\subsubsection{Dynamic JSSPs}

While the above works focus only on static JSSPs, various works have been proposed to handle DyJSSP variants. These scheduling problems involve uncertainties such as random job arrivals, machine breakdowns, and changing job orders. Various works have been proposed that utilize the size-agnostic nature of graphs to solve new DyJSSPs with different problem sizes, which are subject to dynamic changes in online environments. 

\cite{yang2023combining} address a DyJSSP variant with random job arrivals using a deep double Q-network (DDQN) to develop a policy for selecting dispatching rules. They combine a disjunctive graph representation with a specially designed reward space to minimize makespan by comparing operation finish times to optimal solutions. The action space includes only traditional dispatching rules, such as FIFO and SPT, while the node embeddings are learned through a GCN layer. The proposed method's effectiveness is demonstrated by comparing it to general dispatching rules and a DRL-based approach that learns from static state features rather than a graph configuration. Similarly, \cite{zeng2022hybrid} tackles the DyJSSP with machine breakdowns and changing job orders to learn to select dispatching rules from the action set. They model the DyJSSP as an MDP, where the state is defined by disjunctive graphs, and propose self‐attention as the embedding operator and use a double dueling DQN with prioritized replay and noisy networks (D3QPN). The work shows that it outperforms priority dispatching rules and a genetic algorithm. It also highlights the effectiveness of the D3QPN for this learning task by comparing it with other DRL algorithms. In a continuation of their work, \cite{zeng2023you} propose a framework called “Train Once For All” (TOFA). TOFA combines an attention mechanism and spatial pyramid pooling to compress disjunctive graph representations into fixed-length feature vectors and applies D3QPN for learning to select dispatching rules. Experiments highlight its ability to outperform dispatching rules.

Opposed to these works, \cite{liu2023dynamic} presents a DRL-based method using a custom GNN model to learn dispatching rules. This work extends the graph representation of L2D to handle DyJSSPs, utilizing operations from only arrived jobs to construct a disjunctive graph representation to handle stochastic job arrivals and random machine breakdowns. Information on the processing times, degree of completion of operations, and machine availability is updated over time in the dynamic environment. PPO is employed to optimize the agent's policy, and extensive testing in dynamic and static environments shows it outperforms dispatching rules and a genetic algorithm. Besides this, \cite{su2023evolution} proposes to learn a similar policy using an evolution strategies (ES) based algorithm, which enhances stability and robustness compared to conventional DRL approaches. The proposed framework is used to solve DyJSSPs with machine breakdowns and stochastic processing times of operations and is found to converge within the same number of episodes without relying on expensive backpropagation and a value function. This can significantly reduce the required training time. It has also been demonstrated to be more robust in training, showing less dependence on hyperparameters and initial starting points.

Besides DyJSSP, \cite{lei2023large}, in an extension of the work from \citet{lei2022end}, introduces a hierarchical reinforcement learning (HRL) framework to address dynamic \textit{flexible} job shop scheduling problems (DyFJSSP) with new job arrivals and flexible machine assignments. This framework includes a higher-level layer agent and two lower-level layer agents. The higher-level agent breaks the DyFJSSP into a static FJSSP by deciding when to release online arrived jobs, while the lower-level agents handle the machine allocation and the dispatching of operations. The MDP for the higher-level agent is defined by the number of cached jobs and the variance of completion times of all machines after rescheduling. A disjunctive graph is used as a state representation for the lower-level job operation selection agent, and the machine assignment agent uses a state representation that consists of the completion time and processing time of the operation to be assigned to a machine. A GIN encodes the disjunctive graph state representation for the job operation selection agent, and an MLP is used to encode the higher-level agent and the machine assignment task state representations. A DDQN trains the higher-level agent on when to release jobs, and the two lower-level agents are trained with a multi-PPO. This extension of the PPO algorithm learns two joint policies to solve the static sub-problem, which involves two tasks with two actors sharing a single joint state value function (i.e., the critic). Correspondingly, the method is evaluated using a simulation model that simulates the dynamic production environment where jobs arrive with a Poisson distribution. Results indicate that the method outperforms dispatching rules and an expensive ant colony optimization (ACO).
\cite{zhang2023deep} also investigate the DyFJSSP, considering variable processing times. The work uses the PPO algorithm to train how to dispatch jobs to different machines based on the disjunctive graph state space configuration. To consider the variability of processing times, the work trains and tests agents on a number of Brandimarte and Hurink instances in which completion times are generated at random. The results demonstrate that the proposed method outperforms a genetic algorithm (GA) and an ACO algorithm in most static cases and single scheduling rules in dynamic environments.
\par
\cite{zhang2023dynamic} tackle another variant, being the DyFJSSP with insufficient transportation resources. They design a heterogeneous graph model with machine and AGV nodes and extract the node features through a three-stage embedding method. PPO is employed for policy network training, demonstrating enhanced performance on Ham's benchmarks \citep{Ham2020}, which are an extension of Hurink's benchmark instances.

\subsubsection{Distributed JSSPs}
The DiJSSP extends the classical JSSP by incorporating multiple identical factories, each with a set of machines. A collection of independent jobs, each consisting of a series of operations, must be scheduled across these machines. Once assigned to a factory, a job cannot be reallocated.

To address this problem, \cite{huang2023novel, huang2024end} introduce a novel priority dispatch rule generation method for the DiJSSP, wherein jobs and machines are distributed across multiple factories. The study treats the DiJSSP as an MDP, using disjunctive graph structures to represent jobs and machines within individual factories and then combining these into a unified representation. A factory assignment rule is devised to pre-allocate jobs to factories by sorting all jobs in ascending order of their total processing times and assigning a first set of jobs to separate factories. The remaining jobs are sequentially assigned to the factory with the smallest total processing time, balancing the workload across all factories. The scheduling decisions are made using the PPO algorithm. To ensure robustness and scalability, the proposed approach is validated across multiple instances, demonstrating competitive results and consistent stability. Afterwards, \citet{huang2024hierarchical} propose a multi-action deep reinforcement learning (MDRL) method, where they extend the disjunctive graph of JSSPs to represent the real-time arrivals of jobs and different factories. The GIN is used to extract embeddings of partial schedules, which are then fed into two MLPs to calculate probabilities of selecting eligible jobs and factories. The machine selection in each factory is decided by the earliest available machine. The PPO is customized by two actor-critic components to train both policies for job and factory selection.

\begin{table*}[t]
\centering
\caption{\textbf{GNN-based methods for FSPs and the variants.}}
\vskip 0.1in
\resizebox{\linewidth}{!}{
\begin{tabular}{@{}c|c|c|c|c|c|c@{}}
\toprule
\midrule
Method                             & Year   & Problem & GNN Task    & Training Algorithm     & Graph Type    & GNN Type             \\ 
\midrule\midrule
\citet{kwon2021matrix}                             & 2021 & HFSP    & Dispatching Rule Generation & REINFORCE                     & Bipartite Graph     & GAT                      \\ \midrule
\citet{ni2021multi}                               & 2021  & HFSP    & Dispatching Rule Generation & PPO                           & Multiple Graphs & GCN \\ \midrule
\citet{Choo2022}                               & 2022  & HFSP    & Learning Improvement Heuristic & REINFORCE                          & Bipartite Graph  & GAT \\ \midrule
\citet{dong2022minimizing}                                & 2022 & PFSP & Dispatching Rule Generation & REINFORCE & Complete Graph & GIN \\ \midrule
\citet{zhou2023reinforcement}                               & 2023  & PFSP    & Dispatching Rule Generation & PPO                          & Disjunctive Graph & GIN \\ \midrule
\citet{li2024learning}                               & 2024  & PFSP    & Imitating NEH Solutions & Supervised Learning                           & k-NN Graph & Gated GCN \\ \midrule
\citet{zhao2024application}                               & 2024  & HFSP    & Dispatching Rule Generation & PPO                           & Heterogeneous Graph & HGNN \\ \midrule
\bottomrule
\end{tabular}}
\label{table:4}
\end{table*}

\subsection{FSPs with the Variants}
The FSP and its variants, such as HFSP and PFSP, are a class of scheduling problems that are similar to JSSPs. The GNN-based methods developed for solving JSSPs can be naturally extended to the FSP domain. 

\citet{dong2022minimizing} regard the solving process of PFSPs as a sequence-to-sequence generation problem, given that the scheduling order of jobs on each machine should be the same. Based on the pointer network~\citep{vinyals2015pointer}, they employ an LSTM network that takes as input a random sequence of jobs in the encoder and outputs the optimized sequence in the decoder. However, they claim that the sequence of jobs in the encoder is meaningless and should not affect the optimal sequence in the decoder. To overcome this issue, they replace the LSTM-based encoder with a GIN, which represents jobs on a complete graph and then derives their embeddings for attention computation in the decoding process. The REINFORCE algorithm is used to adjust the generative probability of job sequences to minimize late work. The improved iterative greedy algorithm is further used to improve solutions obtained by the neural network. Similarly, \citet{pan2021deep} propose a LSTM-based neural network, PFSPNet, and enhance the solution by Nawaz Enscore Ham (NEH) algorithm~\citep{nawaz1983heuristic}. Instead of GNNs, they use a one-dimensional convolution layer to advance the job embeddings in the encoder. 

\citet{li2024learning} come up with a graph representation where nodes signify jobs that are featured by the processing time on each machine. They further sparsify the fully connected graph into a k-nearest neighbors graph (k-NNG). The gated GCN~\citep{joshi2019efficient} is adopted to extract job embeddings. To construct the solution (i.e., the sequence of jobs), a context embedding, including the average job embedding, the first job embedding, and the previous job embedding, interacts with embeddings of unscheduled jobs to compute their probabilities of selection. Supervised learning is used to train the gated GCN to imitate the solutions derived from the NEH algorithm. Their method performs significantly better than the DRL method in~\citep{pan2021deep}.

\citet{ni2021multi} introduce a multi-graph attributed reinforcement learning-based optimization (MGRO) method, in which the Gantt charts of schedules at each stage are transformed into multiple graphs. The graph embeddings are obtained by a GCN to predict search operators to improve the schedules iteratively. This method outperforms traditional methods such as NEH and iterated greedy (IG)~\citep{ruiz2007simple}. 

\citet{zhao2024application} employ heterogeneous graphs to represent the scheduling status in the HFSP so that the machines and jobs can be more differentiable with more informative embeddings. A heterogeneous graph neural network (HGNN) with an attention mechanism is built to capture the embeddings, which are then used to predict the machine-job pair to determine the assignment of jobs to machines. 

\citet{kwon2021matrix} represent the machines and jobs on a bipartite graph, with edges weighted by the processing time of jobs on each machine in the HFSP. The GAT-based MatNet is designed to pass messages between the job nodes and machine nodes. At each stage of the HFSP, the node embeddings of machines interact with embeddings of jobs by the attention mechanism to iteratively select jobs on each machine. The REINFORCE algorithm is used to optimize the policy. \citet{Choo2022} propose an improvement method for combinatorial optimization, similar to EAS~\citep{hottung2021efficient} and COMPASS~\citep{Chalumeau2023}. Instead of the JSSP, they evaluate their method using MatNet~\citep{kwon2021matrix} on the HFSP. The approach called simulation-guided beam search (SGBS), combines beam search with simulation. By alternating expansion, simulation, and pruning phases, SGBS uses trained policies to search for promising schedules. Their experiments demonstrate that SGBS, particularly when combined with EAS, considerably improves the scheduling performance. \citet{zhou2023reinforcement} represents a PFSP instance with the scheduling status by a disjunctive graph. A GIN is used to capture the graph representation and is fed into an MLP to predict the action (i.e., the job selection). By training with the PPO algorithm, the GIN-based model gains better performance than NEH and several metaheuristics. 

Generally, the application of GNNs in FSPs evolves naturally from their use in JSSPs, given the similar definitions. GNNs are commonly employed to extract embeddings for jobs and machines, which are then used to predict the probabilities of selecting jobs or job-machine pairs at each step of the scheduling process.

\section{Discussion}

This section summarizes the main findings of our review. Before delving into the specific characteristics of each scheduling variant, we first address several recurring challenges that reoccur across all subtypes of scheduling problems. These include issues such as the lack of consensus on representation choices, the absence of standardized benchmarking, and concerns related to evaluation metrics, scalability, and adaptability. 

\subsection{Recurring Challenges:}

\begin{itemize}
\item \textbf{Representation Consensus:} For all problem variants—JSSPs, FJSSPs, DyJSSPs, DiJSSPs, and FSPs—there is no agreement on the optimal graph representation for learning. While disjunctive, heterogeneous, and bipartite graphs are commonly used, the lack of comparative studies prevents definitive conclusions about the most suitable format for various contexts. Furthermore, it's plausible that there may be a better methodology for approaching these problems than graph representation. Alternative data structures or modelling techniques offer more efficient or insightful solutions. However, this area remains largely unexplored and could benefit from further research to ascertain the most effective approaches.

\item \textbf{Absence of Standardized Benchmarking:} The lack of standardized benchmarking across different works poses another challenge. Many works use different problem instances, including synthetic or use-case-specific datasets, making it difficult to directly compare the performance of various approaches. Commonly used benchmarks in these studies are often outdated and fail to reflect the scale and complexity of contemporary industrial problems. This discrepancy hinders accurate assessments of these methods' real-world applicability.

\item \textbf{Lack of Standardized Evaluation Metrics:} Evaluation metrics differ between studies, with some focusing on optimality gaps in terms of makespan, while others consider one or more other objectives. This inconsistency in evaluation metrics adds to the difficulty of comparing methods across different problem variants.

\item \textbf{Scalability Issues:} 
While some methods achieve impressive results, there is limited research into how well these methods scale as problem size and complexity increase. Few studies address the trade-offs between the quality of the solutions and computational efficiency for large instances. More work is needed to evaluate the scalability of GNN-based methods and their ability to handle real-world, large-scale scheduling problems.
\end{itemize}

\subsection{Problem-Specific Observations:}

\begin{itemize}
\item \textbf{Regarding JSSPs:} 1) Most existing works train policies acting as dispatching rules, which sequentially add jobs to the schedule until a full plan is created. Recently, a few works have aimed at guiding search or improvement heuristics instead, offering a new promising research direction. 2) The disjunctive graph representation is most commonly used, while several other works propose different graph representations. There is no consensus or comparison on the most effective representation of JSSP instances for graph learning. 
3) Current works mostly deploy prevalent GNN architectures, such as GIN, GAT, or other basic message passing structures. However, recently, several works have proposed other architectures leveraging Transformers. As for the RL algorithms, most works adopt the widely used PPO or REINFORCE to train the policy network. 4) It is unclear which methods are the most effective, as authors often did not test with the same benchmark instances. 
Most works use part of the commonly known benchmarks such as ABZ, LA, ORB, and TA, while others create synthetic or use-case-specific datasets. 
5) Regarding performance, 
the state-of-the-art learned constructive policies with greedy inference achieve optimality gaps ranging from 6 to 20\% on the benchmark datasets. For larger instances (e.g. $100\times20$), learned policies can outperform the CP-SAT solver with a time limit of one hour. The improvement heuristic policy by \citet{zhang2024deep} achieves better performance, achieving (near-)optimal results on many instances at the cost of more runtime than 
constructive policies. 6) Several works have aimed to improve the inference performance over greedy or random sampling. EAS, for example, focuses on adjusting the final layer of the GNN agent to adapt to specific problem instances. SGBS and COMPASS propose promising search techniques to optimally use the probabilistic outputs of the GNN agents to obtain better solutions.

\item \textbf{Regarding FJSSPs:} 1) Most research uses DRL and GNNs to learn PDRs rather than simply selecting or applying traditional PDRs. This allows models to automatically discover and learn scheduling rules from data, adapting to different scheduling scenarios and objectives with high generality. 2) For the FJSSP, using heterogeneous graphs for the representation is a common method. This structure handles graphs with different nodes and edges, where nodes can represent machines or jobs, and edges can represent different relationships between them. However, there is no consensus on the best representation for learning methods. 3) The current work primarily employs popular GNN architectures such as GIN, GCN, MPGN, or DDN, and DRL algorithms mostly utilize PPO, Actor-Critic, or other variants. 4) Various FJSSP benchmark instances, such as Brandimarte and Hurink data, are primarily used as part of the experiments, with some employing self-generated synthetic instances. 5) The use of multi-agent methods enhances system flexibility and scalability, enabling better adaptation to environmental changes and providing more flexible scheduling strategies. 6) Some work aims to improve the performance of GNNs, such as incorporating attention mechanisms. GATs efficiently represent complex operations and relationships between machines in graphs, effectively extracting useful features through attention mechanisms.


\item \textbf{Regarding DyJSSPs and DiJSSPs:} 1) Most existing works extend approaches developed for basic JSSPs to dynamic and distributed environments. For DyJSSPs, the current research focuses on both selecting and generating PDRs, whereas the limited research on DiJSSPs primarily concentrates on rule generation. 2) For both DyJSSPs and DiJSSPs, there is a consensus in the analyzed works to use disjunctive graph representations. However, there is a research gap in exploring alternative graph formulations, i.e., to embed multiple factories into a single graph-based formulation. 3) The GIN architecture is most frequently used for generating embeddings, with PPO as the training algorithm. Most works are trained in an offline environment and later deployed to solve problems with dynamic configurations. There is a research gap in developing learning methods that more effectively adapt to the dynamic nature of these problems. 4) Current works are typically benchmarked against commonly used PDRs. However, there is limited comparison between different learning methods due to variations in problem dynamics and the lack of a standardized simulation model or publically available standardized instances with dynamic events.

\item \textbf{Regarding FSPs:} 1) The GNN applications in FSPs are not as popular as in JSSPs. Compared to FSPs, JSSPs are featured by an additional precedence constraint, which requires that an operation cannot begin until the preceding operation of the same job is completed. This constraint enables JSSPs to encompass a wider array of real-case scenarios, highlighting the unique challenges and opportunities presented by JSSPs. 2) The GNN-based methods use different types of graphs to represent FSPs, such as bipartite, complete, and multiple graphs. 
There is a lack of comparative studies to determine the most preferred graph type.
3) The existing works on FSPs generally employ the commonly used GNN types, such as GCN, GAT, and GIN, 
to capture the embeddings of jobs or machines. Additional MLPs take in the embeddings for subsequent job or machine selection. A special case is that GCN can also be used to learn the representation of the Gantt chart. 4) Similar to JSSPs, most works on FSPs still used REINFORCE or PPO algorithms to train the neural networks. 5) The randomly generated data and public benchmark datasets such as TA and VRF~\citep{vallada2015new,fernandez2020design} are often tested in the experiments, and NEH is often taken as the main baseline. However, a comprehensive comparison of GNN-based methods is still missing.
\end{itemize}

According to the observations, most differences in approaches for each problem variant lie in graph representations, GNNs, and DRL algorithms. There is no one single technique that is best for all problem variants. In general, specialized techniques for each specific variant can bring benefits in delivering more effective solutions. This survey aims to summarize existing methods in JSSPs and present their differences for readers' reference. We also highlighted more comprehensive and fair benchmarking is important to figure out the state-of-the-art techniques for each problem variant, so as to provide more advice and guidance for algorithm development.

\section{Potential Research Directions}




According to our review of existing research, we propose the following potential research topics, which we think are paramount and warrant further attention.

\begin{itemize}
\item \textbf{Generic foundation model:} 
The methods reviewed by this survey highlight the design of GNNs to effectively tackle different JSSPs, especially depending on RL training approaches. 
However, most existing methods train a GNN for each specific problem.
Similar to recent advancements of foundation models in routing problems~\citep{liu2024multi,zhou2024mvmoe}, training one single GNN model that is effective for solving a group of JSSP variants is more favorable, as it offers better generality and reduces heavy training overheads. 
While current GNN-based approaches often adopt a disjunctive graph representation, this static graph structure exhibits limitations in representing certain JSSP variants. For example, it is still inadequate to represent dynamic scheduling scenarios, such as the dynamic JSSP and dynamic FJSSP, where new jobs may come on the fly and in a stochastic fashion. 
Therefore, developing a more effective disjunctive graph structure with a broader and generalized representation is an interesting future research direction. 
The more general graph structure facilitates the development of more unified and versatile GNN models for shop scheduling. Besides the widely-used disjunctive graph representation, another potential direction is to cast the problems into a new formulation, such as the task of sequence generation~\citep{shutler2003priority}, so that a unified model can be proposed for various problem variants. 

\item \textbf{Performance:} Despite advancements, the performance of existing GNN-based methods for static JSSPs is still far from optimal and outperformed by the top-performing heuristic or meta-heuristic methods. 
Hence, designing powerful learning-based algorithms to surpass state-of-the-art heuristics is still an open research problem. This goal can be approached from two main angles. Existing methods often execute the policy learning in a constructive or improving heuristic for JSSPs, with the policy parameterized by GNNs and trained by DRL algorithms. In general, the GNNs’ role is to effectively capture and encode the complex, high-dimensional state space of the JSSP into low-dimensional embeddings that can be used to approximate a (favorable) rule in the heuristic (e.g., the dispatching rule in the constructive heuristic).  GNNs also
help in learning the interdependencies and constraints of the problem, thereby enhancing the decision-making capabilities of heuristics. In contrast, the role of RL is modeling the interactions between the GNN (i.e., agent) and the heuristic (i.e., environment) as a MDP process and updating the GNN to optimize the objective. Accordingly, we can improve the current learning-based methods from three main angles. Firstly, developing 
tailored GNNs that excel in learning effective policies for shop scheduling problems is crucial for improving performance. However, an optimal model cannot be achieved without employing a more effective problem representation. Therefore, exploring enhanced representation learning techniques alongside problem formulation strategies is recommended. Secondly, most existing GNN-based methods attempt to learn dispatch rules. Extending GNNs to work with more advanced solvers (e.g., meta-heuristic methods or exact solvers) is promising to further boost the performance. Thirdly, there is a need to explore more advanced DRL algorithms or machine learning paradigms to enhance performance further or improve generalization. For example, more advanced DRL algorithms should be investigated to further narrow the optimal gap for JSSPs. 
To improve generalization, meta-learning~\citep{huisman2021survey} is a promising paradigm and has already shown effectiveness in the routing domain~\citep{zhou2023towards}. Its combination with scheduling is a promising direction for future research.

\item \textbf{Practicality:} Scheduling problems in real-world manufacturing systems typically entail many practical constraints such as sequence-dependent setup times, machine availability windows, and temporal constraints between operations. In addition, these problems often involve multiple conflicting objectives, such as minimizing makespan while maximizing throughput. However, the existing research predominantly focuses on single-objective optimization, with complex constraints being infrequently addressed. An important future research topic is to extend existing GNN-based approaches to tackling multi-objective and multi-constraint shop scheduling problems, which is essential for advancing the practical applicability of learning-based methods in real-world settings.

\item \textbf{Robustness:} Given the stochasticity of parameters in JSSPs and FSPs, such as stochastic processing times and machine availability, it is important to entail robust solutions when GNNs are used to learn the scheduling policy. Current research focuses more on achieving solutions with average performance on a group of instances in terms of makespan. However, producing reliable and trustworthy solutions should be the first priority. In this case, extending existing GNNs to involve stochasticity is promising by coupling with stochastic neural networks (SNN) or Bayesian neural networks (BNN). In addition, risk-aware objectives such as the worst-case Conditional Value-at-Risk (CVaR) could be used to enhance the robustness of the learned solutions.

\item \textbf{Explainability:} While most works attempt to learn PDRs by GNNs, achieving desirable results to some extent, there is still no research on explaining the learned policies. This gap hinders the practical application of GNNs compared to hand-crafted PDRs.
Explainable GNN models or post-hoc explainability techniques are highly demanded to make the policy learning process, or the learned solutions more understandable and trustworthy to developers or users. For example, the general GNNExplainer models or surrogate models such as decision trees and linear models~\citep{ying2019gnnexplainer,yuan2022explainability,saeed2023explainable} can be employed to approximate behaviors of learned PDRs so as to make them more understandable.

\item \textbf{Benchmarking:} Benchmarking is essential for evaluating the effectiveness and efficiency of scheduling algorithms. Despite the availability of benchmarks for classical scheduling problems, there is a notable lack of benchmarks for many variants, especially for dynamic job shop scheduling problems (DyJSSPs) and problems with practical constraints. This gap hinders comprehensive evaluation and fair comparison of algorithms under consistent conditions. Currently, we are aware of only one benchmarking environment, \citet{reijnen2023job}, that supports the evaluation of scheduling algorithms under static and dynamic conditions. Future research should focus on developing more unified benchmark instances and environments, enabling fair comparisons of proposed methods.
\end{itemize}

\section{Conclusion}
This paper provides a comprehensive survey of the burgeoning field of applying graph neural networks (GNNs) to address JSSPs and FSPs. By systematically reviewing the existing literature, we have outlined various graph representations employed for different types of JSSPs and FSPs, along with the diverse GNN architectures. Additionally, we have analyzed the current methodologies, emphasizing the crucial components of GNN applications such as graph representation, GNN types, task specifics, and RL algorithms. Through this survey, we have identified several strengths and weaknesses of utilizing GNNs for JSSPs and FSPs. While GNNs offer promising opportunities for improved scheduling solutions, there are challenges related to scalability, interpretability, and computational complexity that need to be addressed by the community. Furthermore, the choice of RL algorithms influences the effectiveness of GNN-based approaches.

Looking ahead, this survey points towards exciting avenues for future research. Opportunities exist for the development of novel GNN architectures tailored specifically to JSSPs and FSPs, as well as the exploration of hybrid approaches combining GNNs with other optimization techniques. Additionally, developing generic foundation models and addressing the scalability, practicality, robustness, and interpretability challenges of GNNs in scheduling contexts remain important areas for further investigation. Overall, this survey serves as a valuable resource for researchers and practitioners interested in leveraging GNNs to tackle the complexities of JSSPs and FSPs, aiming to advance scheduling theory and practice.








\newpage
\bibliographystyle{elsarticle-harv}
\bibliography{main}
\end{document}